\documentclass[a4paper, oneside, twocolumn, notitlepage, 10pt]{extarticle_ecoc}
\usepackage{ecoc}
\usepackage{ecoc}
\usepackage{subfigure}
\usepackage{siunitx}
\usepackage{tikz}
\usepackage{float}
\begin{document}
\selectlanguage{english}    


\title{\vspace{-5pt} 
Advanced Equalization in 112 Gb/s Upstream PON Using a Novel Fourier Convolution-based Network
\\}%

\author{
    Chen Shao\textsuperscript{(1)}, 
    Elias Giacoumidis \textsuperscript{(2)},
    Patrick Matalla \textsuperscript{(3)}, 
    Jialei Li \textsuperscript{(1)},
    Shi Li \textsuperscript{(2)},  \\
    Sebastian Randel \textsuperscript{(3)},
    Andre Richter \textsuperscript{(2)},
    Michael F\"arber \textsuperscript{(1)},
    Tobias K\"afer \textsuperscript{(1)}
}

\maketitle                  


\begin{strip}
 \begin{author_descr}

   \textsuperscript{(1)} Karlsruhe Institute of Technology, AIFB,
   \textcolor{blue}{\uline{chen.shao2@kit.edu}}

   \textsuperscript{(2)} VPIphotonics GmbH,
   \textcolor{blue}{\uline{elias.giacoumidis@vpiphotonics.com}} 

   \textsuperscript{(3)} Karlsruhe Institute of Technology, IPQ, Karlsruhe, Germany
   \textcolor{blue}{\uline{sebastian.randel@kit.edu}}
 \end{author_descr}
\end{strip}

\setstretch{1.1}
\renewcommand\footnotemark{}
\renewcommand\footnoterule{}


\begin{strip}
  \begin{ecoc_abstract}
    We experimentally demonstrate a novel, low-complexity {Fourier Convolution-based Network (FConvNet)} based equalizer for 112 Gb/s upstream PAM4-PON. At a BER of $\sim$ $5\times10^{-3}$, FConvNet enhances the receiver sensitivity by 2 and 1 dB compared to a 51-tap Sato equalizer and benchmark machine learning algorithms, respectively. \textcopyright2024 The Author(s)
  \end{ecoc_abstract}
\end{strip}

\section{Introduction}
\label{sec:intro}
The demand for higher data rates in passive optical networks (PONs) has been increasing due to the data traffic growth \cite{kanedaOFC2020,kiglis2021}. While 50G standard for PONs using intensity modulation and direct detection (IMDD) with OOK has been recently agreed upon, research is now focused on 100 Gb/s PAM4 \cite{Borkowski2020WorldsFF} which is more prone to nonlinearities.
The dynamic range of packets received at the optical line terminal (OLT) in upstream PONs, due to differential loss, poses a challenge implementing burst-mode trans-impedance amplifiers (TIAs) at 50 GBaud PAM4 \cite{fariba2023,WangOFC2024,gurneECOC2022}. 
Additionally, these PON links need to adhere to the 29 dB power budget of legacy PON systems \cite{fariba2023}. To address this, semiconductor optical amplifiers (SOAs) can be used as pre-amplifiers in the OLT receiver to enhance receiver sensitivity. However, operating the SOA with a constant bias leads to degradation in high-power packets due to gain saturation-induced patterning effects \cite{fariba2023}. The distortions caused by the nonlinear behavior of the SOA typically require more advanced DSP compared to traditional feed-forward equalizers (FFEs), such as deep neural networks (DNNs) \cite{MatallaOFC2024}. However, to guarantee energy-efficient PONs, it is essential to have lower complexity compared to DNNs. 

A frequency-calibrated sampling convolutional and interaction network (FC-SCINet) equalizer was simulated for a downstream 100G PON with a path loss of 28.7 dB \cite{chen2024novel}. At 5 km, FC-SCINet improved the bit-error-ratio (BER) by 88.87\% compared to FFE and a 2-layer DNN with 10.57\% lower complexity. FC-SCINet enables effective time-series modeling with intricate temporal dynamics and fine-tuning of signal spectral attributes.

In this paper, we present a novel machine learning algorithm (MLA) for equalization, named Fourier attention-based convolutional neural network (FConvNet), and compare it with FC-SCINet in an experimental 112 Gb/s PAM4 upstream PON setup. FConvNet is based on TimesNet \cite{wu2022timesnet}, incorporating multi-periodicity in time series analysis by capturing recurring cycles within the data. In upstream PONs, the SOA combined with direct-detection produces interference to the signal, introducing multi-periodicity through nonlinear effects, crosstalk, and reflections. Converting the conventional 1D time series into 2D tensors captures intra-period and inter-period variations, enhancing the analysis of cycle relationships. A 2D representation, considering time and frequency domains as dimensions, integrates temporal and spectral information for a comprehensive interpretation. This method converts the initial 1D time series into organized 2D tensors, making it easier to process with 2D kernels. It can also handle multivariate time series by reshaping all variables \cite{wu2022timesnet}. FConvNet also incorporates a CNN (i.e., ConvNet) with a Decomposition layer inspired by FC-SCInet \cite{chen2024novel}, which helps reduce complexity and accelerates training efficiency. FConvNet improves chromatic dispersion (CD) tolerance at 2.2 km compared to a 51-tap Sato equalizer, achieving a 2 dB received optical power (ROP) improvement and a 1 dB ROP enhancement compared to DNN, FC-SCINet, and CNN at a BER of \textasciitilde $5\times10^{-3}$. FConvNet reduces the complexity by 79\% and 83.4\% compared to DNN and CNN, respectively.
\begin{figure*}[t!]
  \centering
  \captionsetup{font=footnotesize} 
  \subfigure[]
  {
    \includegraphics[width=0.65\textwidth, height=5cm]{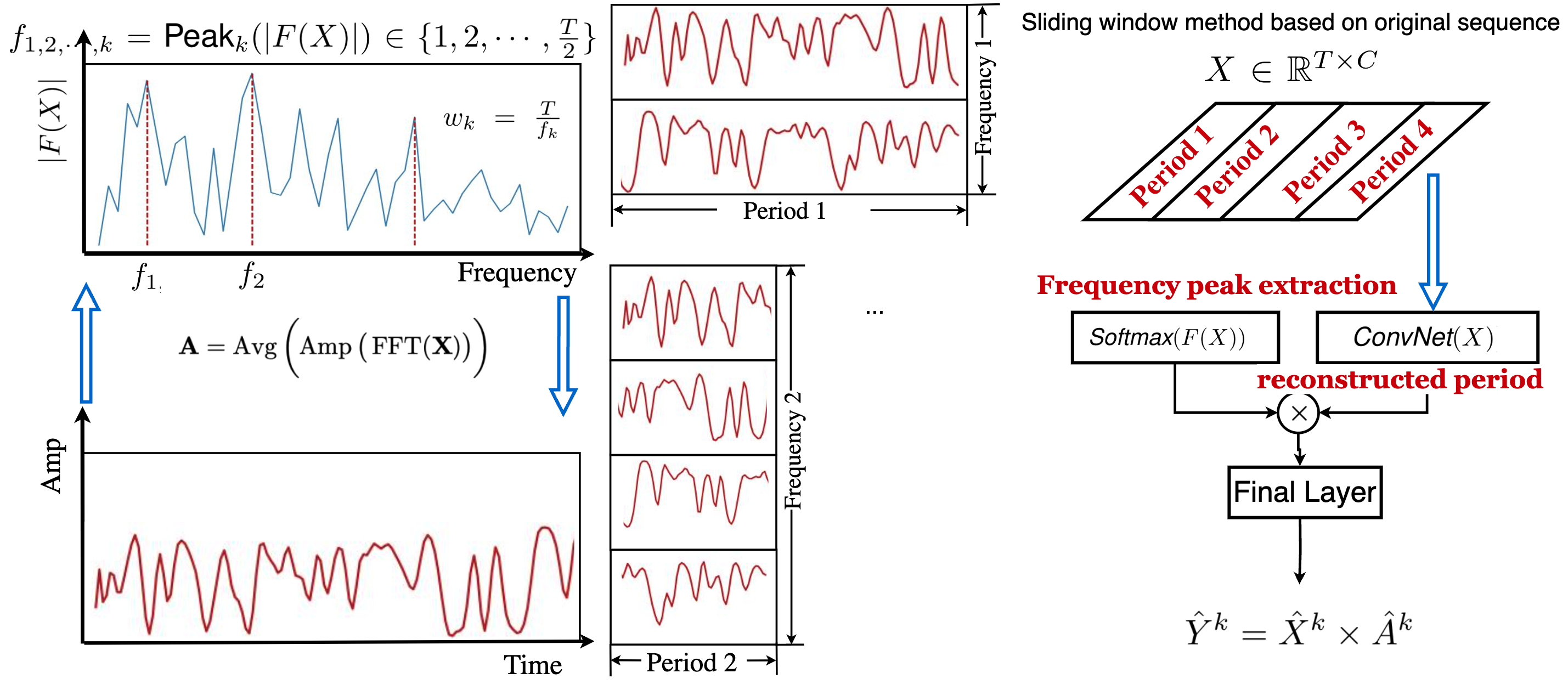}
    \label{subfig:1a}
    }
  \subfigure[]
  {
    \includegraphics[width=0.31\textwidth, height=5cm]{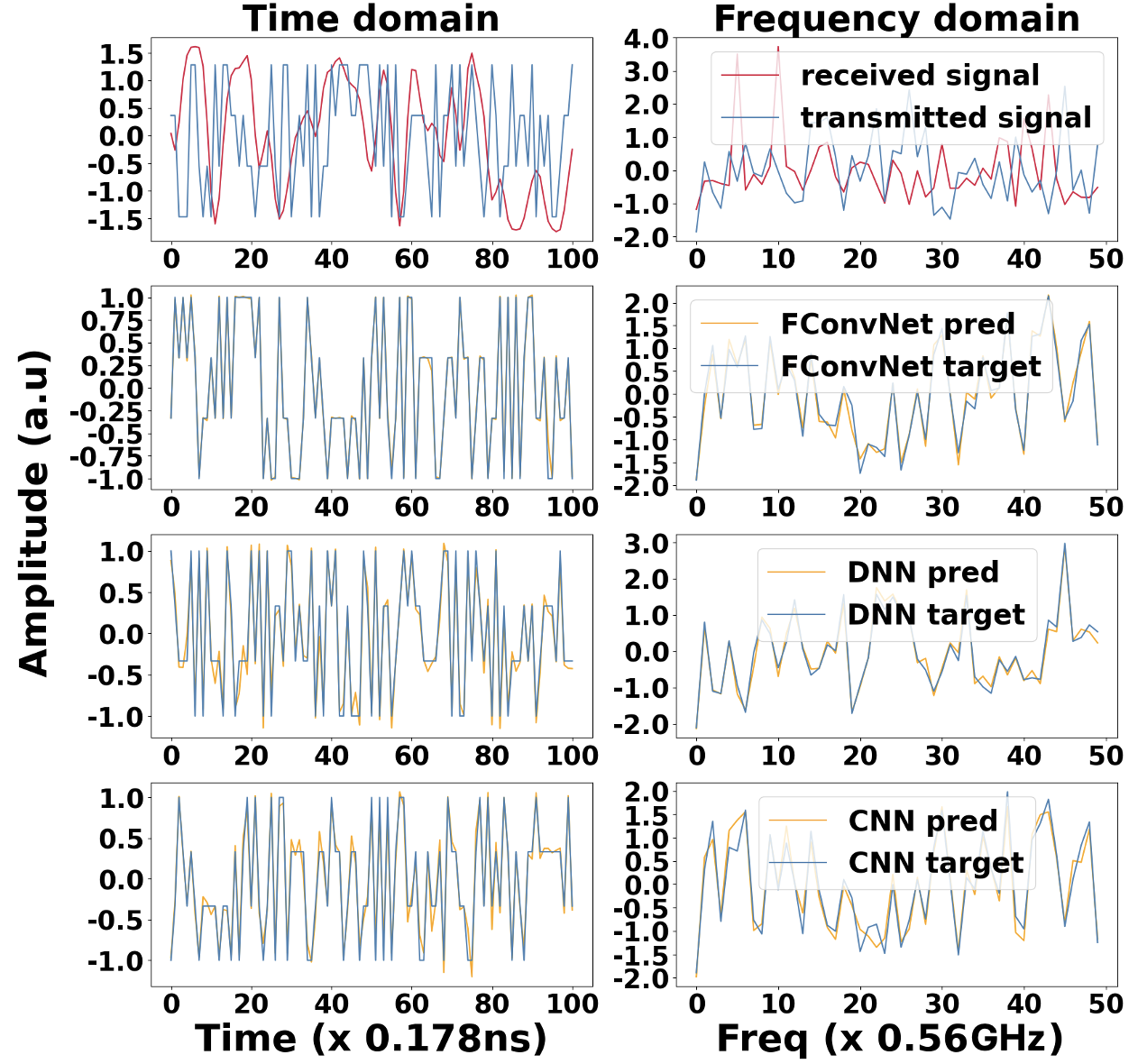}
    \label{subfig:1b}
    } 
  \caption{(a) Block diagram illustrating the novel FConvNet equalizer. (b) Time/Frequency domain representation of 100 transmitted (target)/received consecutive samples at -5 dBm received power, showing the prediction capability of FConvNet, DNN, and CNN.}  
  \label{fig:Fig1}
\end{figure*}
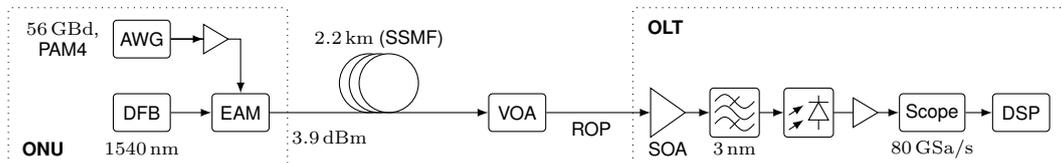
\begin{figure*}[b]
    \centering
    \begin{tikzpicture}
[scale=1.3,
block/.style={draw,align=center,rounded corners = 0.05cm, minimum width=1.0cm,minimum height=0.5cm},
block2/.style={draw,align=center,rounded corners = 0.05cm, minimum width=0.75cm,minimum height=0.5cm}]

\scriptsize
    \node[block2,fill=white] at (0,0) (awg1) {AWG};
    \node[align=center] at (-0.8,0) {\SI{56}{GBd},\\PAM4};
    \node[block2,fill=white] at (1,-0.75) (eam1) {EAM};
    \node[block2,fill=white] at (0,-0.75) (dfb1) {DFB};
    \node[align=center] at (0,-1.12) {\SI{1540}{nm}};
    
    \draw[-latex] (dfb1) -- (eam1);
    \draw[-latex] (awg1) -- (1,0) -- (eam1);
    \draw[fill=white] (0.625,-0.15) -- (0.625,0.15) -- (0.875,0) -- cycle;
    \draw (awg1) -- (0.625,0);

    \draw[-latex] (awg1) -- (0.625,0);

    \node[align=center] at (2.4,0) {\SI{2.2}{km} (SSMF)}; 
    \node[align=center] at (1.9,-1) {\SI{3.9}{dBm}};
    \draw[fill=white] (2.3,-0.45) circle (0.3);
    \draw[fill=white] (2.4,-0.45) circle (0.3);
    \draw[fill=white] (2.5,-0.45) circle (0.3);

    \draw [rounded corners=0.05cm, dotted] (-1.35,-1.32) rectangle (1.475,0.32); 
    \node[align=center] at (-1,-1.12) {\textbf{ONU}}; 

    \node[block2,fill=white] at (3.8,-0.75) (voa) {VOA};
    \draw[-latex] (eam1) -- (voa);
    \node[align=center] at (4.55,-0.92) {ROP};

    \draw[fill=white] (5.15,-1) -- (5.15,-0.5) -- (5.5,-0.75) -- cycle;
    \node[align=center] at (5.325,-1.12) {SOA};
    \draw[-latex] (voa) -- (5.15,-0.75);
    
    \draw [rounded corners=0.05cm] (5.75,-1) rectangle (6.25,-0.5);
    \draw (5.8,-0.75) sin (5.9,-0.68) cos (6,-0.75) sin (6.1,-0.82) cos (6.2,-0.75);
    \draw (5.8,-0.635) sin (5.9,-0.565) cos (6,-0.635) sin (6.1,-0.705) cos (6.2,-0.635);
    \draw (5.8,-0.865) sin (5.9,-0.795) cos (6,-0.865) sin (6.1,-0.935) cos (6.2,-0.865);
    \draw (5.95,-0.685) -- (6.05,-0.585);
    \draw (5.95,-0.915) -- (6.05,-0.815);
    \draw[-latex] (5.5,-0.75) -- (5.75,-0.75);
    \node[align=center] at (6,-1.12) {\SI{3}{nm}};

    \draw [rounded corners=0.05cm] (6.5,-1) rectangle (7,-0.5);
    \draw (6.85,-0.925) -- (6.85,-0.575);
    \draw[fill=white] (6.85,-0.645) -- (6.75,-0.825) -- (6.95,-0.825) --  cycle;
    \draw (6.75,-0.645) -- (6.95,-0.645);
    \draw[-latex] (6.55,-0.775) -- (6.7,-0.675);
    \draw[-latex] (6.55,-0.925) -- (6.7,-0.825);
    \draw[-latex] (6.25,-0.75) -- (6.5,-0.75);

    \draw (7,-0.75) -- (7.2,-0.75);

    \draw (7.2,-0.9) -- (7.2,-0.6) -- (7.45,-0.75) --  cycle;
    \node[block2] at (8,-0.75) (scope) {Scope};
    \draw[-latex] (7.45,-0.75) -- (scope);
    \node[align=center] at (8,-1.12) {\SI{80}{GSa/s}};

    \node[block2] at (8.9,-0.75) (dsp) {DSP};
    \draw[-latex] (scope) -- (dsp);

    \draw [rounded corners=0.05cm, dotted] (4.97,-1.32) rectangle (9.35,0.32); 
    \node[align=center] at (5.3,0.12) {\textbf{OLT}}; 

\end{tikzpicture}	
    \caption{Experimental setup for the \SI{112}{Gb/s} (\SI{56}{GBaud}) PON upstream using a standard single-mode fiber (SSMF) in the C-band.} 
    \label{fig:experimental_setup}
\end{figure*}

\section{Proposed Equalizer}
\label{sec:model}
The FConvNet consists of three modules: The \textbf{Signal Partition}, \textbf{ConvNet}, and \textbf{Reconstruction}. FConvNet processed 13 random non-repeating number sequences (RNS) for each ROP from the OLT, each with a length of $T$ from $C$ independent recorded trials, represented as $X \in \mathbb{R}^{T \times C}$. The equalized sequence is denoted as $Y \in \mathbb{R}^{T \times C}$. Initially, we assess the sliding time window approach with window sizes ($ws$) on $X \in \mathbb{R}^{ws \times C}$. Then, a single-layer convolution neuron (kernel size $s_{k1}$ = 3) captures the initial temporal features. Similarly to FC-SCINet \cite{chen2024novel}, we utilize the Decomposition layer to enhance high-fluctuating components. 

Inspired by multi-resolution analysis methods \cite{192463}, FConvNet performs fine-scale equalization operations on the time-frequency bins with the highest energy concentration and combines their signals using weighting factors based on energy density. Essentially, we view the total channel distortion as the result of multiplying a spectrum-based distortion $\sigma_1(W_1\vert F(X)\vert)$ with the transformed time domain values $\sigma_2(W_2X)$. The $\vert F(X)\vert$ denotes the amplitude spectrum of $X$. The $\sigma(WX)$ represents the neural network with parameter $W$ performed on $X$. $\sigma_1$ and $\sigma_2$ are approximated by Softmax and ConvNet as shown in Fig. \ref{subfig:1a}.
\vspace{-5pt}
\begin{equation}
\label{equ:equ1}
\vspace{-6pt}
Y = \textit{Softmax}(F(X))\times(\textit{ConvNet}(X))
\end{equation}

\noindent \textit{Signal Partition}: 
We select the peak-k amplitude values $f_{1,2, \cdots,k} = \text{Peak}_k(\vert F(X)\vert) \in \{1, 2, \cdots, \frac{T}{2}\}$ $f_2$ with corresponding window sizes $ w_k = \frac{T}{f_k}$. Based on $k$ window sizes, we partition the $X$ into k sub-series with length $\frac{T}{f_k} \in \{T/2, \cdots, T\}$. After zero padding, we reshape them into a 2D matrix $X^{k}$ corresponding to $w_k$. The signal with the same length shares the same minimal time resolution $w_k$. The spectral component $A^k=\vert F(X^k)\vert$ is followed by a Softmax layer to learn the nonlinear interference in frequency domain $\hat{A}^k$.

\noindent \textit{ConvNet}: The ConvNet receives the reshaped input $X^{k}$. Afterwards, the interference is estimated in the time domain denoted as  $\sigma_2$, utilizing the Inception block \cite{Szegedy2014GoingDW}, which is a component used in CNNs allowing for the integration of multiple different kernel sizes and receptive fields within a single layer. The Inception block consists of concatenated convolutions with kernel sizes of 1, 3, and 5. It uses a Gaussian error Linear Unit (GeLU), which serves as a smoother nonlinear activation function in comparison to ReLU. The resulting output is denoted as $\hat{X}^{k}$.

\noindent \textit{Reconstruction}: Followed by Eq. \ref{equ:equ1}, we multiply $\hat{X}^{k}$ with $\hat{A^k}$ to reshape them into the desired form, resulting in  $\hat{Y}^k$, as shown in Eq. \ref{equ:equ2}.
\begin{equation}
    \hat{Y}^{k} = \hat{X}^{k} \times \hat{A}^k
\label{equ:equ2}
\end{equation}

\section{Experimental Setup and Integrated DSP}
The experimental 112 Gb/s upstream PON setup is depicted in Fig.~\ref{fig:experimental_setup}. In the ONU, we used a Keysight USPA DAC3 to generate a \SI{56}{GBaud} NRZ PAM signal with 4 levels (PAM4) and a sequence length of $2^{16}$ symbols. The electric drive signal was amplified to a peak-to-peak voltage of \SI{2}{V} to achieve sufficient optical modulation amplitude at the low-cost electro-absorption modulator (EAM), resulting in a transmitted power of \textasciitilde \SI{3.9}{dBm}. A distributed feedback (DFB) laser provided the optical carrier at a wavelength of \textasciitilde \SI{1540}{nm}. After a 2.2 km transmission, the signal was attenuated by a variable optical attenuator (VOA) to set a specific ROP. The  OLT receiver consisted of an SOA with a \SI{3}{nm}-bandpass filter to account for the higher SNR requirements of PAM4 compared to OOK. A \SI{40}{GHz}-PIN photodiode coupled with a conventional amplifier was utilized due to the unavailability of an avalanche photodiode with TIA in our laboratory. It is worth noting the latter option could have potentially enhanced the receiver sensitivity. Finally, the electrical signal was captured by a \SI{33}{GHz}-real-time oscilloscope at \SI{80}{GSa/s} and resampled to twofold oversampling for blind feed-forward clock recovery before equalization. 

We addressed potential performance overestimation caused by bit pattern recognition using a bit stream based on RNS. For comparison, we tested a 2-layer DNN, CNN, and FC-SCINet \cite{chen2024novel}. To find the optimal DNN architecture, we employed a range of 4 to 128 neurons. Note that incorporating an additional layer resulted in negligible performance benefits. For CNN, a single linear-layer architecture with 48 neurons was employed, as well as double linear-layer architectures with 2048 and 256 neurons, respectively, serving as detectors. During the training phase for all MLAs, L2 regularization was used, while the mean square error function was utilized as the loss function. 15\% of the data was used for testing and 10\% for validation. For comparison with a linear equalizer, we used a decision-directed FFE with 21/51 finite impulse response taps using the Sato algorithm \cite{Matalla} to adapt the filter coefficients.
\section{Results and Discussion}
\begin{figure*}[t]
  \centering
  \captionsetup{font=footnotesize} 
  \subfigure[]
  {
    \includegraphics[width=0.22\textwidth]{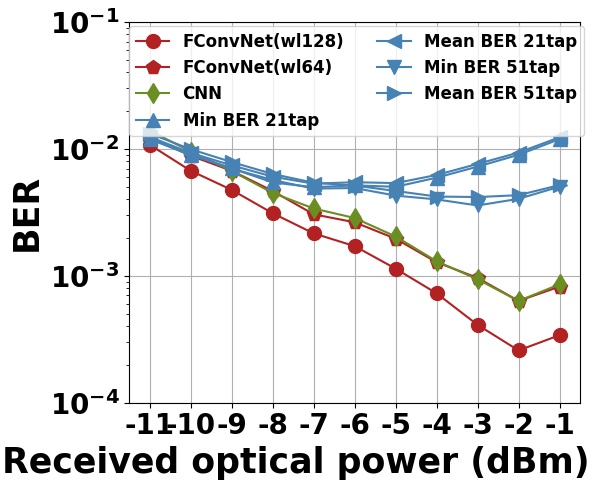}
    \label{subfig:3a}
    }
      \subfigure[]
  {
    \includegraphics[width=0.22\textwidth]{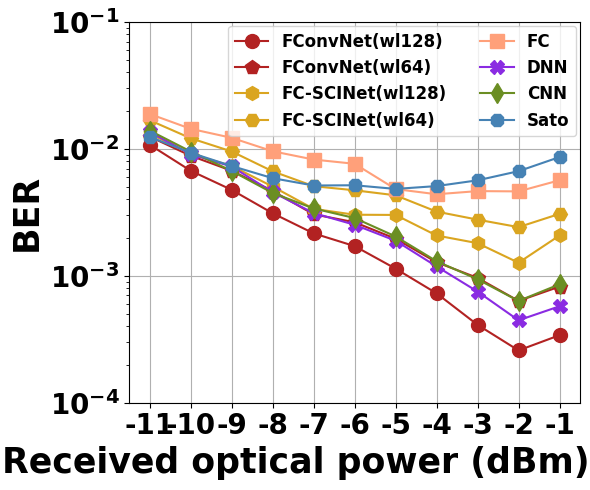}
    \label{subfig:3b}
    }
  \subfigure[]
  {
    \includegraphics[width=0.4\textwidth]{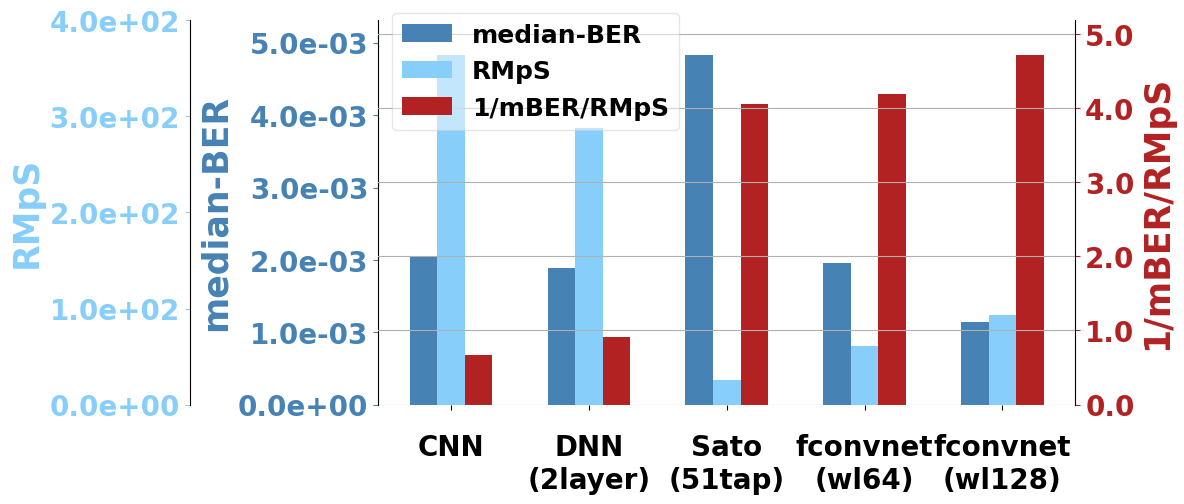}
    \label{subfig:3c}
    }
  \caption{BER vs. received optical power for FConvNet [window length (wl=64, 128)] compared to: (a) Sato (min./mean BER) and CNN; (b) FC-SCINet [wl=64, 128], FC, DNN, CNN and Sato (min. BER, 51 taps). (c) Complexity comparison of different models.}  
  \label{fig:Fig2}
\end{figure*}

FConvNet performance was compared with that of an optimized 2-layer DNN (2048, 256 neurons), a CNN, and the FC-SCINet reported in \cite{chen2024novel} all at 1 SpS, while Sato equalization at 2 SpS, using 2$^{16}$ symbols. All MLAs exhibited similar performance at 2 SpS; however, for complexity reduction purposes, we used 1 SpS. FC without SCInet was also compared, enabling only frequency-components calibration using the Decomposition layer \cite{chen2024novel}. Evaluation was conducted by directly counting the BER through Monte-Carlo across various ROPs.

As depicted in Fig. \ref{subfig:3a} and Fig. \ref{subfig:3b} at a BER of \textasciitilde $5\times10^{-3}$, FConvNet demonstrates a 2 dB improvement in ROP compared to a 51-tap Sato equalizer (considering the min. BER). Moreover, 1 dB enhancement in ROP is achieved compared to a 2-layer DNN, FC-SCINet and CNN. Note, the BER was improved when increasing the window length (wl128) in FConvNet. Fig. \ref{subfig:1b} showcases the time/frequency domain representation of 100 consecutive transmitted/received samples at -5 dBm ROP, highlighting FConvNet has superior predictive capabilities in time domain compared to DNN/CNN due to pulse narrowing.
 
Finally, we compared the complexity of FConvNet (64/128wl) with the aforementioned MLAs. The complexity analysis encompassed key metrics including Real Multiplications per Symbol (RMpS), median BER (mBER), and the composite metric $\frac{1}{\text{mBER} \times \text{RMpS}}$ (BER-C), which serves as an indicator of the performance-complexity trade-off (higher is better), as shown in Fig. \ref{subfig:3c}. FConvNet's complexity can be calculated as $\text{RMpS}_{\text{FConvNet}} = n_{k_{1}} d_{\text{model}} (n_{s} - n_{k_{1}} + 1) + 2 \cdot n^{2}_{s} + 2(d_{\text{I}} n_{k_{2}}  d_{\text{II}})(n_{s} - n_{k_{2}} + 1)$, where we only consider the number of the multiplications. $n_{s}$ is the sequence length of the input, $d_{\text{I}}$ $d_{\text{II}}$ and $n_{k_{1}}$, $n_{k_{2}}$  are the dimension and kernel sizes of two Inception blocks utilized in ConvNet, respectively. 
FConvNet achieves 79.0\% reduction in RMpS and 85.7\% in BER-C compared to DNN, and 83.4\% and 80.5\% reductions compared to CNN. It also lowers the BER-C by 13.9\% compared to Sato equalization.

\label{sec:conclusion}
\section{Conclusion}
A novel FConvNet equalizer was experimentally demonstrated for a 112 Gb/s upstream PAM4-PON. At a BER of $\sim$ $5\times10^{-3}$, it enhanced the ROP by 2 and 1 dB compared to a 51-tap Sato equalizer and benchmark MLAs, respectively. FConvNet reduced complexity by 79\% and 83.4\% compared to DNN and CNN, respectively.

\newpage
\section{Acknowledgements}
This work was funded by the Federal Ministry of Education and Research (BMBF) (KIGLIS: 16KIS1228, 16KIS1230). 
\printbibliography
\vspace{-4mm}
\end{document}